\documentclass[11pt]{article}
\usepackage[a4paper,margin=25mm]{geometry}
\usepackage[T1]{fontenc}
\usepackage{lmodern,microtype,amsmath,amssymb,booktabs,tabularx,graphicx}
\usepackage{xcolor,tikz}
\usetikzlibrary{arrows.meta,positioning}
\usepackage[numbers,sort&compress]{natbib}
\usepackage[colorlinks=true,urlcolor=blue!55!black,citecolor=blue!55!black,linkcolor=blue!55!black]{hyperref}
\usepackage{enumitem}
\setlist{nosep}
\title{Kinora: A Versioned Workflow for Text-to-Motion Generation, Robot Retargeting, and Interactive Validation}
\author{Fangcheng Jin\\University of Chinese Academy of Sciences\\\texttt{jinfangcheng2023@ia.ac.cn}}
\date{Technical report --- November 9, 2026}
\begin{document}
\maketitle
\begin{abstract}
Motion generation tools produce candidate trajectories, but interactive robot authoring also requires recoverable execution, traceable edits, explicit geometry, and model-specific validation. We present Kinora, a system that connects text-to-motion generation, a motion library, local agent tools, robot retargeting, and trajectory refinement through versioned artifacts. The current release uses an independently specified 52-joint authoring rig and procedural skeleton previews, separating the distributed human geometry from externally installed generation models and robot assets. Human-motion edits preserve frames outside a selected interval, while robot edits create immutable candidates whose reports govern acceptance and export. We evaluate server 0.5.1 and Gadget 0.15.2 with 156 core regression test executions, 49 Gadget executions, four CUDA generation/editing cases, a 30-frame G1 retargeting check, and a 3,600-frame synthetic G1 fixture paired with a suspended negative control. All four CUDA cases passed their numerical gates; the turning splice nevertheless retained a 14.3 cm support-foot drift proxy warning. The G1 positive fixture completed bounded simulation replay, whereas the suspended control was quarantined before replay. These results support selected integration and preservation contracts. They do not establish semantic-quality improvements, population-level feasibility, or hardware safety.
\end{abstract}
\noindent\textbf{Artifacts:} \url{https://github.com/PoirotJin/kinora}

\section{Introduction}
Text-conditioned human motion generation provides a useful starting point for animation and robot motion authoring. A generated trajectory, however, is only one intermediate artifact in a longer process. A user may search for an existing motion, generate a variation, repair a short interval, join two clips, retarget the result to a robot, adjust a joint interactively, and export an accepted version. Each operation introduces a different question: which source was used, which frames changed, which robot model was assumed, and which checks actually ran?

These questions become harder when the workflow spans a desktop application, a remote server, and an external language-model agent. Retried requests can duplicate expensive jobs; an edit can silently replace its source; a report can refer to an older candidate; and a visually plausible preview can be mistaken for a dynamically feasible trajectory. A practical system must preserve the relationship between requests, artifacts, versions, and validation outcomes across these interfaces.

Kinora organizes this workflow around immutable candidates and explicit acceptance. Text-to-motion inference uses Kimodo~\citep{kimodo}; Kinora contributes the surrounding execution and authoring workflow rather than a new generative model. The server records durable jobs, the Gadget exposes account-scoped operations to local agents, and the robot editor associates each candidate with model-specific checks. A successful numerical or physical check is interpreted only within its stated scope.

This report makes three concrete contributions. First, it describes a shared artifact and execution contract for generation, local editing, retargeting, and agent access. Second, it makes the human-geometry contract explicit through an independent numerical rig, motion-specific geometry checks, and a candidate-based robot refinement workflow with bounded simulation replay. Third, it documents reproducible release-level checks and diagnostic cases, including a failed physical control and a remaining splice-quality warning. The evaluation is intentionally limited to engineering evidence available for this release; no comparison with state-of-the-art generation methods or population-level success rate is claimed.

\section{Related Work and Scope}
\paragraph{Generation and retrieval.}
Human Motion Diffusion Model (MDM)~\citep{mdm} studies diffusion-based human motion synthesis, while Kimodo~\citep{kimodo} addresses controllable generation at scale. Kinora integrates Kimodo as its active generation backend. TMR~\citep{tmr} learns a joint text--motion representation for retrieval; Kinora includes a semantic-library path that requires separately installed retrieval assets. That path was not part of the runtime evaluation reported here. The distinction between an upstream model's capabilities and the behavior of this particular integration is essential: results reported for an upstream model are not Kinora measurements.

\paragraph{Robot simulation and tooling.}
MuJoCo~\citep{mujoco} supplies the physics engine used in the robot checks. The broader robot tooling ecosystem includes Holosoma~\citep{holosoma}, and Kinora's distribution documents external runtime and model dependencies. Our focus is the authoring and acceptance workflow, not a new retargeting objective, dynamics solver, or learned controller. Simulation-based checks remain conditional on model parameters, contact assumptions, and the controller used during replay.

\paragraph{System scope.}
The release includes desktop installers, server source, and the local Gadget. Generation, retrieval, email delivery, and particular robot models require additional assets or services. The active generation path is Kimodo; distributed human geometry is defined by the independent rig described below. Older motion files require the explicit compatibility checks rather than an assumption of universal support. Public availability of the server archive should not be interpreted as a grant of unrestricted rights to separately distributed model weights or third-party components.

\section{System Architecture and Artifact Contracts}
\subsection{A shared workflow across interfaces}
Kinora exposes a common server workflow to desktop clients and Gadget. The Gadget provides a command-line and Model Context Protocol interface for a locally configured agent. The local agent can prepare an exact English motion description and request a server operation without forwarding its own model-provider credentials to the server. Server authentication and account permissions remain separate from the agent's language-model access.

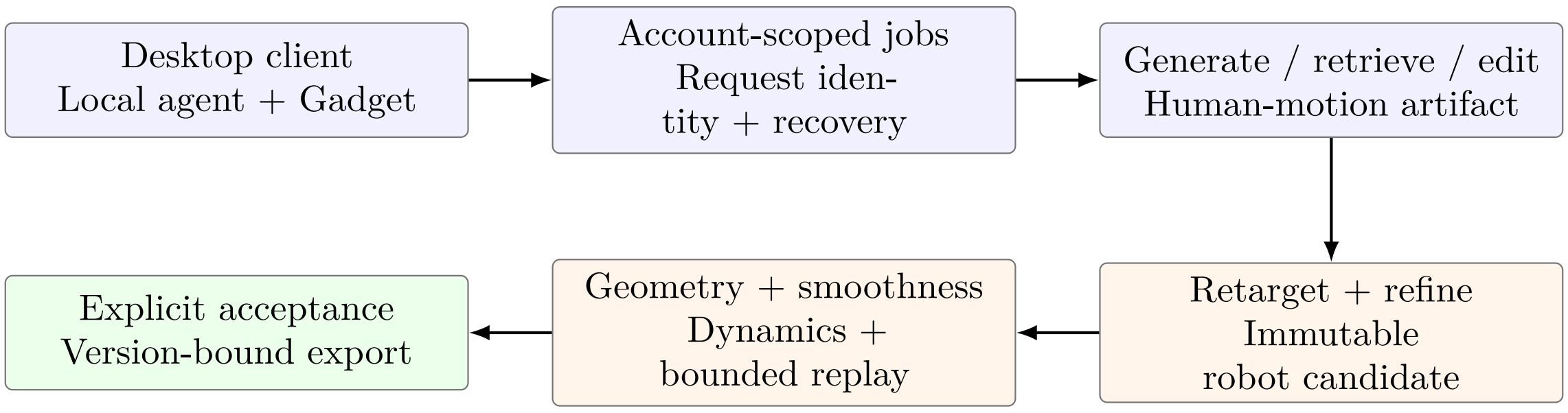
\begin{figure}[t]
\centering
\begin{tikzpicture}[node distance=8mm, box/.style={draw=black!55,rounded corners=2pt,align=center,font=\small,minimum height=11mm,text width=42mm},arr/.style={-{Latex[length=2mm]},thick}]
\node[box,fill=blue!6] (ui) {Desktop client\\Local agent + Gadget};
\node[box,right=of ui,fill=blue!6] (jobs) {Account-scoped jobs\\Request identity + recovery};
\node[box,right=of jobs,fill=blue!6] (human) {Generate / retrieve / edit\\Human-motion artifact};
\node[box,below=12mm of human,fill=orange!8] (robot) {Retarget + refine\\Immutable robot candidate};
\node[box,left=of robot,fill=orange!8] (checks) {Geometry + smoothness\\Dynamics + bounded replay};
\node[box,left=of checks,fill=green!8] (accept) {Explicit acceptance\\Version-bound export};
\draw[arr] (ui)--(jobs);\draw[arr] (jobs)--(human);\draw[arr] (human)--(robot);\draw[arr] (robot)--(checks);\draw[arr] (checks)--(accept);
\end{tikzpicture}
\caption{Kinora's artifact workflow. Validation belongs to a particular candidate and robot configuration. A failed or unavailable check does not become a successful result through export or a different client interface.}
\label{fig:pipeline}
\end{figure}

Durable job records use SQLite and account-scoped state. Stable request identifiers and payload hashes support idempotent handling of retries and recovery after interruption. This design gives a client a way to recover the status of an existing request rather than treating a disconnected response as proof that the operation never ran. The regression suite exercises protocol and recovery behavior, but this report does not claim distributed exactly-once execution under arbitrary failures.

\subsection{Versioned motion artifacts}
A human-motion artifact contains time-indexed motion arrays and metadata describing generation or editing. New generator outputs additionally embed rest-joint coordinates and a parent hierarchy; older four-field numeric imports remain a separate compatibility path. The current generation interface supports 1--120 s at 30 frames per second and a 52-joint SMPL-H-compatible NPZ representation. NPZ and BVH exports support downstream use. Source identity, edit selection, generation settings, and validation reports provide provenance for derived candidates.

For a source sequence $X=(x_0,\ldots,x_{T-1})$ and a selected interval $I=[a,b)$, local editing constructs a candidate $Y$ with the contract
\begin{equation}
Y_t=X_t\quad\text{for all }t\notin I.
\label{eq:exterior}
\end{equation}
The implementation checks exact equality of the relevant stored time-indexed arrays outside the selection. This contract prevents an apparently local operation from drifting the remainder of the clip. It does not by itself establish continuity at the two edit boundaries or semantic correctness inside the interval; those properties require separate checks.

Robot editing similarly preserves the source and creates a new candidate. Undo and redo navigate versions rather than relying on destructive in-place changes to the only copy. Acceptance is a deliberate state transition associated with the validated candidate. A stale report, a read-only session, or an unaccepted candidate cannot be used to bypass the tested export restrictions.

\subsection{An explicit human-geometry contract}
A portable authoring artifact needs a defined skeleton in addition to pose channels. Kinora Human52 is a fixed numerical design in metres, with a Y-up, positive-Z-forward T pose, 22 body joints, and 30 finger joints. Its offsets are specified in the distributed source rather than loaded from a downloaded human skeleton, skinned body, or texture asset. The channel ordering is a compatibility interface; the rig is not a learned body model or an anatomically calibrated reconstruction. External Kimodo weights and its source representation remain upstream dependencies, and robot geometry is installed separately.

The adapter transfers named body-joint rotations from the upstream representation into this target hierarchy and then evaluates forward kinematics (FK). Let $\pi(j)$ be joint $j$'s parent, $b_j$ its rest offset, $R_{t,j}$ its local rotation, and $G_{t,j}$ its global rotation. For a non-root joint,
\begin{equation}
G_{t,j}=G_{t,\pi(j)}R_{t,j},\qquad
p_{t,j}=p_{t,\pi(j)}+G_{t,\pi(j)}b_j.
\label{eq:rigfk}
\end{equation}
The rest coordinates, hierarchy, joint names, and their digests define a reproducible target. Procedural previews draw these joints and bones; they do not require a textured human mesh. This geometry choice makes deployment explicit but does not improve the upstream generator by itself. Generated fingers remain fixed, and no dexterous hand-retargeting result is claimed.

Motion-specific geometry is preserved when compatible older numeric files do not embed a rig. Given stored points and rotations, the implementation estimates each rest offset by a componentwise median over frames,
\begin{equation}
\widehat b_j=\operatorname{median}_{t}\left[
G_{t,\pi(j)}^{\mathsf T}(p_{t,j}-p_{t,\pi(j)})\right].
\label{eq:recover}
\end{equation}
It reconstructs the relative joint coordinates and requires a maximum componentwise discrepancy of at most 5 mm before using that geometry. Inconsistent files are rejected rather than silently assigned the new rig's bone lengths. This check does not uniquely identify the absolute rest-root origin of every historical format: ambiguous grounding conventions require migration review. Original files are not rewritten by recovery. New embedded geometry avoids this ambiguity, while the separate four-field import/export path does not preserve every optional metadata field.

\section{Generation, Repair, and Splicing}
\subsection{Generation and local repair}
The generation adapter translates a text request, duration, seed, and optional constraints into the upstream Kimodo execution path. The server retains resulting metadata and exposes progress through its job interface. Missing model configuration causes an explicit refusal instead of a synthetic success response. A two-second CUDA preflight provides a small deployment check before longer jobs.

Local repair uses an inference window around the requested interval and merges the candidate into the source while preserving the exterior in Eq.~\eqref{eq:exterior}. The inference-window length can differ from the final artifact length; it should not be interpreted as the exported clip duration. Boundary checks report translational and rotational discontinuities independently of the generator's textual instruction.

\subsection{Staged transitions between clips}
Two-clip splicing must reconcile both endpoint pose and root displacement. Kinora constructs a bridge with four stages: finish the source action and stand, walk toward the target region, slow and stand, and enter the target action. The staged description and root-following constraints provide explicit structure for the generation request. They do not guarantee correct foot contacts or semantic adherence.

The implementation checks the assembled sequence and the edit seams at different levels. At the edit seams, recorded limits are 0.20 m for the maximum joint-position step, 120 m/s$^2$ for the joint-acceleration proxy, and $\pi/6$ radians for the rotation step. Sequence-level limits are 0.25 m, 150 m/s$^2$, and $\pi/4$ radians, respectively. These are engineering rejection bounds, not perceptual-quality targets. The two layers use different limits and should not be pooled into a single quality score.

For a joint position $p_t$ sampled at interval $\Delta t$, the finite-difference quantities have the form
\begin{equation}
 d_t=\|p_t-p_{t-1}\|_2,\qquad
 a_t=\|p_{t+1}-2p_t+p_{t-1}\|_2/\Delta t^2.
\end{equation}
Reports also expose root continuity and gait proxies. In particular, estimated support-foot drift is a diagnostic derived from the human trajectory, not a verified sole-contact measurement. A splice can pass the broad numerical bounds and still have a visible or physically problematic foot-motion pattern. The evaluation preserves this distinction instead of discarding warnings when a case passes.

\section{Robot Refinement and Conditional Validation}
\subsection{Model-aware editing}
Retargeting produces a robot-specific candidate that can be inspected with mesh-based forward kinematics. The editor supports joint manipulation, inverse-kinematics interactions, keyframes, and version history. Custom robot profiles bind mappings and limits to a particular model configuration so that a report is not silently interpreted against another robot. The release includes these paths, although the independent runtime case in this report uses G1 only.

Geometric checks and smoothness checks answer different questions. Geometric validation concerns the candidate's consistency with its robot representation; temporal checks constrain velocity, acceleration, and jerk. The editing policy includes joint bounds of 4 rad/s, 40 rad/s$^2$, and 800 rad/s$^3$, and root-translation bounds of 0.5 m/s, 5 m/s$^2$, and 100 m/s$^3$. These are refinement-policy settings, not manufacturer-certified operating envelopes. Missing required physical parameters prevents a successful dynamics claim.

\subsection{Inverse dynamics and bounded replay}
The dynamics stage evaluates whether a candidate requires unexplained root support or exceeds declared actuator and velocity limits. Let $m$ denote model mass and $g$ gravitational acceleration. The residual checks normalize root force and torque as
\begin{equation}
\rho_f=\|f_{\mathrm{root}}\|/(mg),\qquad
\rho_\tau=\|\tau_{\mathrm{root}}\|/(mg\ell),\quad \ell=1\,\mathrm{m}.
\end{equation}
The tested policy requires each residual ratio to remain at or below 0.01. Actuator utilization and joint-speed utilization are measured relative to the configured limits, with a $10^{-6}$ relative numerical tolerance.

Candidates passing the prerequisite checks undergo continuous MuJoCo replay with a torque-bounded inverse-SPD controller. Replay clips actuator torques to their configured ratings and does not use root actuators or reset the state at each reference frame. Tracking checks bound root-position error by 0.02 m, root-rotation error by 0.05 rad, and joint-position error by 0.05 rad. This controller is an evaluation mechanism; it is not a claim about a manufacturer's deployed controller.

The candidate's final status controls the next user action. A ready candidate still requires explicit acceptance before the tested export path permits release of the accepted version. A quarantined candidate remains available for diagnosis but cannot acquire an accepted export simply because some earlier checks succeeded. The physical meaning of a successful report is limited to the tested model, trajectory, controller, and simulation assumptions.

\section{Experimental Evaluation}
\subsection{Questions, environment, and evidence}
The evaluation addresses three engineering questions: whether distribution and interface checks pass in the tested setup, whether selected human-motion operations preserve their edit contracts, and whether robot validation distinguishes a controlled feasible fixture from an unsupported one. It does not estimate average generation quality or compare synthesis algorithms.

Runtime checks used Linux x86\_64, Python 3.10, a 48 GiB NVIDIA GPU, PyTorch 2.5.1 with CUDA 12.4, and MuJoCo 3.4.0. The frontend model tests additionally used Node.js 22.16.0. Evaluation ran in an isolated temporary directory with independent outputs. The fresh archive check used a new extraction and data directory but reused installed Python dependencies; GPU and retargeting tests likewise reused host runtimes and authorized model assets. This is therefore not evidence of a complete model installation on a blank GPU machine.

The released server source is version 0.5.1 and Gadget is version 0.15.2. The CUDA execution used upstream Kimodo commit \texttt{e1c219c99b4b94ba48be1fb6a1cbc64ecee4c954}. The source archive includes a per-file manifest, test scripts, setup instructions, and asset checks. We report stored validation receipts rather than replacing missing measurements with estimates.

\subsection{Distribution and interface checks}
\begin{table}[t]
\centering
\caption{Release checks and their evidential scope. Counts are test executions or inspected assets, not independent samples of real-world reliability.}
\label{tab:release}
\small
\begin{tabularx}{\textwidth}{@{}l r X@{}}
\toprule
Check & Result & Scope \\
\midrule
Core regression & 156 passed & Backend protocol, edits, duration, robot editor/dynamics, account and permission paths. \\
Gadget regression & 49 passed & Local protocol, HTTP permission behavior, and recovery. \\
Pinned external assets & 29 verified & G1 robot files only; content hashes checked. \\
Independent human rig & 7 tests passed & Offline regeneration, FK consistency, recovery, rejection, pivot preservation, export, and splice checks; included in the core count. \\
G1 retargeting & 30 frames & Actual generated-source conversion and finite robot output; not a dynamics certificate. \\
CUDA diagnostic cases & 4 passed & Numerical gates and selected edit invariants; quality warnings retained. \\
Fresh archive startup & Passed & New data directory, account/login, service endpoints, and missing-model refusal. \\
\bottomrule
\end{tabularx}
\end{table}

Table~\ref{tab:release} summarizes the server and Gadget checks. Test modules cover both expected behavior and selected failure paths. Gadget 0.15.2 changes format descriptions and the package version without changing protocol behavior; its 49-test suite was rerun after that patch. A subsequent 16-test HTTP suite and two diagnostic-tool tests passed after terminology cleanup; these overlapping executions are not added to the 156 core count. Some test discovery/import relationships can repeat checks, so 156 and 49 should be read as execution counts rather than unique experimental conditions. Email-related tests use a test substitute; they do not demonstrate successful delivery through an external SMTP provider.

Separate desktop acceptance records cover fresh installation and overwrite upgrade from 0.4.18 to 0.5.0 on Ubuntu 22.04, Ubuntu 24.04, and Windows Server 2025, for six scenarios in total. These records check version, process, desktop-to-server communication, and preservation of settings. They do not constitute a full interactive usability study or verification on every consumer Windows version. This earlier desktop evidence is supplementary and is not presented as a new installation study of desktop 0.5.1.

\subsection{CUDA generation and editing diagnostics}
Four fixed-seed diagnostic cases exercise the CUDA integration. The source request uses seed 210 and generates five seconds of relaxed standing. Repair uses seed 211 to modify frames $[25,125)$ of that 150-frame source, requesting a gentle left-hand lift. Two splice cases use seeds 212 and 213. They join source segments $[0,60)$ and $[90,150)$ through a 240-frame bridge; the target is displaced by 0.4 m in the straight case, and by $(0.2,0.4)$ m with a 30-degree turn in the turning case. Each staged inference window has 270 frames, including surrounding context, with stage lengths 65, 90, 50, and 65 frames.

\begin{table}[t]
\centering
\caption{Measured edit-seam diagnostics. Maxima aggregate the start and end seams within each case. All three edits preserve the exterior exactly. ``Exact cut'' refers to the recorded preservation-through-acceleration flag, not the broad numerical pass condition.}
\label{tab:cuda}
\small
\begin{tabular}{@{}lrrrrl@{}}
\toprule
Case & Step (m) & Accel. (m/s$^2$) & Rotation (rad) & Drift (cm) & Exact cut \\
\midrule
Repair & 0.01023 & 2.086 & 0.0364 & --- & Yes \\
Splice, straight & 0.07118 & 63.888 & 0.3568 & 3.1 & No \\
Splice, turn & 0.01771 & 15.939 & 0.0653 & 14.3 & No \\
\midrule
Seam rejection bound & 0.20000 & 120.000 & 0.5236 & --- & --- \\
\bottomrule
\end{tabular}
\end{table}

All four cases passed their sequence-level checks, and all three edits passed the edit checks and exact exterior preservation. Table~\ref{tab:cuda} reports seam maxima, not the larger internal boundary quantities from the separate sequence validator. Repair additionally preserved the recorded boundary condition through acceleration. The splice cases did not satisfy that exact-cut flag, despite remaining below the numerical rejection bounds.

The splice receipts expose remaining quality limitations. The straight and turning cases have forward-distance fractions of 0.976 and 0.920, respectively; this proxy classifies moving distance using a speed threshold of 0.15 m/s and a forward cosine threshold of 0.5. Their maximum support-foot drift proxies are 0.0307 and 0.1427 m. The straight case records no gait warning, while the turning case records a suspected support-foot drift warning. These are measurements from the current rig and rerun; differences from an earlier release are not a controlled ablation of the geometry change. Sole contact, semantics, robot collision, and robot dynamics remain unverified by these human-motion receipts.

\subsection{Independent-rig integration checks}
Seven regression cases exercise the geometry contract within the core suite. The descriptor test regenerates the bundled coordinates, hierarchy, names, and provenance while network access is blocked and checks byte equality. Further cases check FK consistency of embedded geometry; recovery from a synthetic rig scaled by 1.03 and translated by $(0.02,-0.015,0.01)$ m; rejection of an inconsistent joint perturbation; preservation of a nonzero rest pivot through a turn and BVH export; and unchanged exterior frames when splicing compatible embedded and recovered rigs. The asset-manifest case verifies that the 29 external files are robot assets. These tests cover declared invariants, not all historical motion conventions.

A separate execution feeds the first 30 frames of the current generated standing source through the account-scoped numeric-import retargeting path for the 23-DoF G1. The solver returns a finite $30\times30$ generalized-position array: 23 joint coordinates plus seven free-root coordinates. This confirms that the new human geometry reaches the actual robot solver. It does not establish that the result satisfies the full dynamics export gate, and it is not substituted for the longer controlled robot experiment below.

\subsection{G1 positive fixture and suspended control}
The robot experiment starts from a synthetic 3,600-frame, 30 Hz G1 trajectory containing a balanced posture and arm movement. It tests the refinement and acceptance workflow on a controlled input; it is not a text-generated walking sequence. A negative control translates the trajectory's root upward by 1 m, creating an unsupported suspended motion while preserving its general joint pattern.

\begin{table}[t]
\centering
\caption{G1 validation results. Replay quantities for the suspended control are unmeasured because prerequisite dynamics checks rejected it. Lower values indicate smaller residuals, utilization, or tracking error; they are not cross-model quality scores.}
\label{tab:g1}
\small
\begin{tabular}{@{}lrrr@{}}
\toprule
Quantity & Positive fixture & Suspended control & Limit \\
\midrule
Root-force ratio & $2.15\!\times\!10^{-5}$ & 1.00001 & 0.01 \\
Root-torque ratio & $8.93\!\times\!10^{-5}$ & 0.02052 & 0.01 \\
Inverse-dynamics actuator utilization & 0.14306 & 0.04843 & 1.0 \\
Joint-speed utilization & 0.00102 & 0.00032 & 1.0 \\
Replay actuator utilization & 0.16023 & --- & 1.0 \\
Replay joint-position error (rad) & 0.00137 & --- & 0.05 \\
Replay root-position error (m) & 0.000204 & --- & 0.02 \\
Replay root-rotation error (rad) & 0.000284 & --- & 0.05 \\
Replay integration steps & 61,183 & Not run & --- \\
Final candidate status & Ready & Quarantined & --- \\
\bottomrule
\end{tabular}
\end{table}

The positive candidate completed all 3,599 reference intervals using 17 integration steps per interval, totaling 61,183 steps over approximately 119.967 s of simulated time. Its maximum replay actuator utilization was 0.16023, with no actuator-saturation steps. The measured tracking errors were below the configured limits (Table~\ref{tab:g1}). The ready candidate was explicitly accepted and exported in the integration check.

The suspended control failed the root-residual checks before replay. Its normalized root-force residual was approximately 1.0, well above the 0.01 limit, even though its joint-speed and actuator-utilization quantities were small. This case illustrates why low joint motion or low joint torque alone is insufficient evidence of support feasibility. Zero-valued replay fields in a rejected receipt must not be presented as successful zero-error replay; no replay was run.

The same integration harness checked source preservation, unchanged frames outside the edit interval, undo/redo, explicit acceptance, read-only restrictions, and rejection of stale reports. These observations support the tested state-transition contract. A single synthetic positive fixture and one negative control do not establish that arbitrary retargeted motions will pass, nor that simulation acceptance transfers to a physical robot.

\section{Limitations and Planned Evaluation}
The principal limitation is the narrow experimental sample. Four CUDA cases, a 30-frame retargeting check, and one synthetic robot trajectory with a suspended control are diagnostic examples selected to exercise specific paths. They do not support confidence intervals for deployment success, claims of semantic improvement, or comparisons with competing systems. No controlled latency, throughput, or cost benchmark is reported because the execution environment reused shared host resources.

Motion quality remains incompletely evaluated. The turning-splice drift warning shows that broad continuity gates can admit imperfect contact behavior. Future evaluation should use a fixed prompt suite covering stationary actions, locomotion, turns, transitions, and failure cases, with multiple seeds and predeclared scoring rules. Semantic ratings, foot-contact measures, and preservation checks should be reported separately. Comparing a direct bridge with the staged bridge under matched prompts and seeds would isolate the contribution of the stage structure; that ablation has not yet been run.

Robot feasibility is conditional on the available model. The G1 fixture is deliberately mild, and no hardware experiment was conducted. Contact-model mismatch, inertial uncertainty, actuator dynamics, state-estimation error, and controller differences can invalidate transfer from simulation. A subsequent study should sweep motions and perturbations, report false acceptance and false rejection under a declared reference procedure, and evaluate hardware only through an appropriate independent test process.

Coverage also varies across deployed components. External SMTP, paid model APIs, TMR retrieval assets, all CASBOT models, and a blank-machine installation of every large model were not exercised in the reported runtime validation. The release instructions state these dependencies explicitly. A multi-user study is still needed to evaluate authoring effort, error recovery, and whether versioned acceptance is understandable in practice.

\section{Conclusion}
Kinora connects motion generation, local editing, robot retargeting, and interactive validation through recoverable jobs and versioned candidates. Its central design choice is to make provenance, validation scope, and explicit acceptance part of the workflow rather than infer readiness from a preview or a successful generator call. The available release checks support selected interface and preservation contracts, while the CUDA warnings and suspended G1 control expose the limits of those contracts. The independent rig makes the distributed human-geometry dependency explicit, without eliminating external model or robot-asset requirements. The released artifacts and validation scripts provide a basis for broader motion-quality, usability, and robot-transfer studies.

\appendix
\section{Reproducibility and Artifact Access}
The public repository~\citep{kinora} is a documentation and download hub. The complete server distribution is the named release asset \texttt{kinora-server-0.5.1.zip}, not the repository's automatically generated source-code archive. The server ZIP contains 428 files, including \texttt{SOURCE\_MANIFEST.json}; the manifest records 427 payload entries. Its SHA-256 digest is:
\begin{center}\footnotesize\ttfamily
a92f66c1dac2f51d6c697f92091c2891\\
a979910c659a225bc7aaaad21ffce3a9
\end{center}
The two displayed lines concatenate into a single digest. Desktop installers are version 0.5.1 and the matching Gadget package is 0.15.2. The source release uses the Human52 rig identifier \texttt{kinora-human52-v1}; its nine generation/edit-validation file digests match the CUDA receipt. Other later changes were terminology and packaging updates with targeted checks, so the full core suite is not represented as a bit-identical execution of every final archive file. Third-party licenses and model-access conditions must be followed separately.

The current distribution bundles its human-rig descriptors and procedural preview implementation, with no external human mesh or texture download in the asset manifest. Package inspection and offline regeneration support this dependency statement; they do not establish that the complete system is independent of all third-party models or software.

A minimal reproduction sequence is to extract the named server ZIP, verify the manifest, follow its setup guide, run the public quickstart test, install the required authorized runtime/model assets, and run the full environment doctor. The release documentation enumerates core regression modules. Independent GPU and robot diagnostics are implemented in \texttt{scripts/validate\_kimodo\_edit.py} and \texttt{scripts/validate\_g1\_editor.py}; their command-line help documents output and runtime arguments. The synthetic G1 fixture is distributed under \texttt{scripts/fixtures/}. GPU tests require the appropriate runtime and are not substituted with CPU-only mocks.

For interpretation, report the code and model revisions, selected seeds and prompts, per-stage pass/fail state, warnings, and whether replay actually ran. Preserve rejected cases as evidence. Retries should not silently replace a failed seed with a successful one when estimating quality in future benchmarks. The results in this report are release-validation observations collected on November 9, 2026, with separate earlier desktop-installation records.


\small
\begin{thebibliography}{9}
\bibitem{kimodo} D. Rempe, M. Petrovich, Y. Yuan, H. Zhang, X. B. Peng, Y. Jiang, et al. \emph{Kimodo: Scaling Controllable Human Motion Generation}. arXiv:2603.15546, 2026. \url{https://arxiv.org/abs/2603.15546}.
\bibitem{mdm} G. Tevet, S. Raab, B. Gordon, Y. Shafir, D. Cohen-Or, and A. H. Bermano. \emph{Human Motion Diffusion Model}. arXiv:2209.14916, 2022. \url{https://arxiv.org/abs/2209.14916}.
\bibitem{tmr} M. Petrovich, M. J. Black, and G. Varol. \emph{TMR: Text-to-Motion Retrieval Using Contrastive 3D Human Motion Synthesis}. In ICCV, 2023. \url{https://arxiv.org/abs/2305.00976}.
\bibitem{mujoco} E. Todorov, T. Erez, and Y. Tassa. \emph{MuJoCo: A physics engine for model-based control}. In IROS, pp. 5026--5033, 2012. doi:10.1109/IROS.2012.6386109.
\bibitem{holosoma} Holosoma contributors. \emph{Holosoma}. Software repository, accessed November 2026. \url{https://github.com/amazon-far/holosoma}.
\bibitem{kinora} Fangcheng Jin. \emph{Kinora: Server 0.5.1 and Gadget 0.15.2}. Software distribution and validation documentation, 2026. \url{https://github.com/PoirotJin/kinora/releases/tag/kinora-v0.5.1}.
\end{thebibliography}
\end{document}